\documentclass[9pt,english,british]{extarticle}
\usepackage{mathptmx}

\usepackage[T1]{fontenc}
\usepackage[latin9]{inputenc}
\usepackage{babel}
\usepackage{array}
\usepackage{multirow}
\usepackage{amsmath}
\usepackage{graphicx}
\usepackage[unicode=true,
 bookmarks=true,bookmarksnumbered=false,bookmarksopen=false,
 breaklinks=false,pdfborder={0 0 0},backref=false,colorlinks=false]
 {hyperref}
\hypersetup{pdftitle={Modified SPLICE, Extension to Non-Stereo Data and Run-Time Compensation for Noise Robust Speech Recognition},
 pdfauthor={D. S. Pavan Kumar, N. Vishnu Prasad, Vikas Joshi, S. Umesh},
 pdfsubject={Noise Robust Speech Recognition},
 pdfkeywords={Robust speech recognition, SPLICE, stereo data, feature normalisation, MFCC}}
\usepackage{breakurl}

\makeatletter

\providecommand{\tabularnewline}{\\}

\usepackage{spconf,bbm}
\name{D. S. Pavan Kumar$^1$, N. Vishnu Prasad$^1$, Vikas Joshi$^{1,2}$, S. Umesh$^1$\thanks{This work was supported  under the SERC project funding SR/S3/EECE/050/2013 of Department of Science and Technology, India.}}
\address{$^1$Department of Electrical Engineering, Indian Institute of Technology Madras, India\\$^2$IBM India Research Labs, India}

\@ifundefined{showcaptionsetup}{}{%
 \PassOptionsToPackage{caption=false}{subfig}}
\usepackage{subfig}
\makeatother

\begin{document}

\title{Modified SPLICE and its Extension to Non-Stereo Data for \\Noise
Robust Speech Recognition}
\maketitle
\begin{abstract}
In this paper, a modification to the training process of the popular
SPLICE algorithm has been proposed for noise robust speech recognition.
The modification is based on feature correlations, and enables this
stereo-based algorithm to improve the performance in all noise conditions,
especially in unseen cases. Further, the modified framework is extended
to work for non-stereo datasets where clean and noisy training utterances,
but not stereo counterparts, are required. Finally, an MLLR-based
computationally efficient run-time noise adaptation method in SPLICE
framework has been proposed. The modified SPLICE shows 8.6\% \emph{absolute}
improvement over SPLICE in Test C of Aurora-2 database, and 2.93\%
overall. Non-stereo method shows 10.37\% and 6.93\% \emph{absolute}
improvements over Aurora-2 and Aurora-4 baseline models respectively.
Run-time adaptation shows 9.89\% \emph{absolute} improvement in modified
framework as compared to SPLICE for Test C, and 4.96\% overall w.r.t.
standard MLLR adaptation on HMMs.
\end{abstract}

\section{Introduction}

The goal of robust speech recognition is to build systems that can
work under different noisy environment conditions. Due to the acoustic
mismatch between training and test conditions, the performance degrades
under noisy environments. \emph{Model Adaptation} and \emph{Feature
Compensation} are two classes of techniques that address this problem.
The former methods adapt the trained models to match the environment,
and the latter methods compensate either or both noisy and clean features
so that they have similar characteristics.

Stereo based piece-wise linear compensation for environments (SPLICE)
is a popular and efficient noise robust feature enhancement technique.
It partitions the noisy feature space into $M$ classes, and learns
a linear transformation based noise compensation for each partition
class during training, using stereo data. Any test vector $\mathbf{y}$
is soft-assigned to one or more classes by computing $p\left(m\,|\,\mathbf{y}\right)$
$\left(m=1,2,\ldots,M\right)$, and is compensated by applying the
weighted combination of linear transformations to get the \emph{cleaned}
version $\widehat{\mathbf{x}}$.
\begin{equation}
\widehat{\mathbf{x}}=\sum_{m=1}^{M}p\left(m\,|\,\mathbf{y}\right)\left(\mathbf{A}_{m}\mathbf{y}+\mathbf{b}_{m}\right)\label{eq:splice}
\end{equation}
$\mathbf{A}_{m}$ and $\mathbf{b}_{m}$ are estimated during training
using stereo data. The training noisy vectors $\{\mathbf{y}\}$ are
modelled using a Gaussian mixture model (GMM) $p(\mathbf{y})$ of
$M$ mixtures, and $p\left(m\,|\,\mathbf{y}\right)$ is calculated
for a test vector as a set of posterior probabilities w.r.t the GMM
$p(\mathbf{y})$. Thus the partition class is decided by the mixture
assignments $p\left(m\,|\,\mathbf{y}\right)$.

Over the last decade, techniques such as maximum mutual information
based training \cite{mmisplice}, speaker normalisation \cite{spkrnormsplice},
uncertainty decoding \cite{udsplice} etc. were introduced in SPLICE
framework. There are two disadvantages of SPLICE. The algorithm fails
when the test noise condition is not seen during training. Also, owing
to its requirement of stereo data for training, the usage of the technique
is quite restricted. So there is an interest in addressing these issues.

In a recent work \cite{unseensplice}, an adaptation framework using
Eigen-SPLICE was proposed to address the problems of unseen noise
conditions. The method involves preparation of quasi stereo data using
the noise frames extracted from non-speech portions of the test utterances.
For this, the recognition system is required to have access to some
clean training utterances for performing run-time adaptation.

In \cite{gonzalez}, a stereo-based feature compensation method was
proposed. Clean and noisy feature spaces were partitioned into vector
quantised (VQ) regions. The stereo vector pairs belonging to $i^{th}$
VQ region in clean space and $j^{th}$ VQ region in noisy space are
classified to the $ij^{th}$ sub-region. Transformations based on
Gaussian whitening expression were estimated from every noisy sub-region
to clean sub-region. But it is not always guaranteed to have enough
data to estimate a full transformation matrix from each sub-region
to other.

In this paper, we propose a simple modification based on an assumption
made by SPLICE on the correlation of training stereo data, which improves
the performance in unseen noise conditions. This method \emph{does
not need any} \emph{adaptation data}, in contrast to the recent work
proposed in literature \cite{unseensplice}. We call this method as
modified SPLICE (M-SPLICE). We also extend M-SPLICE to work for datasets
that are not stereo recorded, with minimal performance degradation
as compared to conventional SPLICE. Finally, we use an MLLR based
run-time noise adaptation framework, which is computationally efficient
and achieves better results than MLLR HMM-adaptation. This method
is done on 13 dimensional MFCCs and does not require two-pass Viterbi
decoding, in contrast to conventional MLLR done on 39 dimensions.

The rest of the paper is organised as follows: a review of SPLICE
is given in Section \ref{sec:Review-of-SPLICE}, proposed modification
to SPLICE is presented in Section \ref{sec:Proposed-Modification},
extension to non-stereo datasets is explained in Section \ref{sec:Non-Stereo-Extension},
run-time noise adaptation is described in Section \ref{sec:Additional-Run-time-Adaptation},
experiments and results are presented in Section \ref{sec:Experimental-Setup},
detailed discussion and comparison of existing versus proposed techniques
is given in Section \ref{sec:Discussion} and the paper is concluded
in Section \ref{sec:Conclusion-and-Future} indicating possible future
extensions.

\section{Review of SPLICE}

\label{sec:Review-of-SPLICE}

As discussed in the introduction, SPLICE algorithm makes the following
two assumptions:
\begin{enumerate}
\item The noisy features $\left\{ \mathbf{y}\right\} $ follow a Gaussian
mixture density of $M$ modes
\begin{equation}
p(\mathbf{y})=\sum_{m=1}^{M}P(m)p(\mathbf{y}\,|\, m)=\sum_{m=1}^{M}\pi_{m}\mathcal{N}\left(\mathbf{y}\,;\,\mathbf{\mu}_{y,m},\Sigma_{y,m}\right)\label{eq:ubm}
\end{equation}

\item The conditional density $p(\mathbf{x}\,|\,\mathbf{y},m)$ is the Gaussian
\begin{equation}
p(\mathbf{x}\,|\,\mathbf{y},m)\sim\mathcal{N}\left(\mathbf{x}\,;\,\mathbf{A}_{m}\mathbf{y}+\mathbf{b}_{m},\Sigma_{x,m}\right)\label{eq:assum2}
\end{equation}
where $\{\mathbf{x}\}$ are the clean features. 
\end{enumerate}
Thus, $\mathbf{A}_{m}$ and $\mathbf{b}_{m}$ parameterise the mixture
specific linear transformations on the noisy vector $\mathbf{y}$.
Here $\mathbf{y}$ and $m$ are independent variables, and $\mathbf{x}$
is dependent on them. Estimate of the \emph{cleaned} feature $\widehat{\mathbf{x}}$
can be obtained in MMSE framework as shown in Eq. (\ref{eq:splice}).

The derivation of SPLICE transformations is briefly discussed next.
Let $\mathbf{W}_{m}=\begin{bmatrix}\mathbf{b}_{m} & \mathbf{A}_{m}\end{bmatrix}$
and $\mathbf{y}'=\begin{bmatrix}1 & \mathbf{y}^{T}\end{bmatrix}^{T}$.
Using $N$ independent pairs of stereo training features $\left\{ \left(\mathbf{x}_{n},\mathbf{y}_{n}\right)\right\} $
and maximising the joint log-likelihood 
\begin{equation}
\mathcal{L}=\sum_{n=1}^{N}\log p(\mathbf{x}_{n},\mathbf{y}_{n})=\sum_{n=1}^{N}\sum_{m=1}^{M}\log\left[p(\mathbf{x}_{n}\,|\,\mathbf{y}_{n},m)p(\mathbf{y}_{n}\,|\, m)P(m)\right]\label{eq:likeli}
\end{equation}
yields
\begin{equation}
\mathbf{W}_{m}=\left[\sum_{n=1}^{N}p\left(m\,|\,\mathbf{y}_{n}\right)\mathbf{x}_{n}\mathbf{y}_{n}'^{T}\right]\left[\sum_{n=1}^{N}p\left(m\,|\,\mathbf{y}_{n}\right)\mathbf{y}_{n}'\mathbf{y}_{n}'^{T}\right]^{-1}\label{eq:splicew}
\end{equation}

Alternatively, sub-optimal update rules of separately estimating $\mathbf{b}_{m}$
and $\mathbf{A}_{m}$ can be derived by initially assuming $\mathbf{A}_{m}$
to be identity matrix while estimating $\mathbf{b}_{m}$, and then
using this $\mathbf{b}_{m}$ to estimate $\mathbf{A}_{m}$.

A perfect correlation between $\mathbf{x}$ and $\mathbf{y}$ is assumed,
and the following approximation is used in deriving Eq. (\ref{eq:splicew})
\cite{ssmafify}. 
\begin{equation}
p\left(m\,|\,\mathbf{x}_{n},\mathbf{y}_{n}\right)\approx p\left(m\,|\,\mathbf{x}_{n}\right)\approx p\left(m\,|\,\mathbf{y}_{n}\right)\label{eq:perfcorr}
\end{equation}

Given mixture index $m$, Eq. (\ref{eq:splicew}) can be shown to
give the MMSE estimator of $\widehat{\mathbf{x}}_{m}=\mathbf{A}_{m}\mathbf{y}+\mathbf{b}_{m}$
\cite{lideng}, given by 
\begin{equation}
\widehat{\mathbf{x}}_{m}=\mathbf{\mu}_{x,m}+\Sigma_{xy,m}\Sigma_{y,m}^{-1}\left(\mathbf{y}-\mathbf{\mu}_{y,m}\right)\label{eq:mmsee}
\end{equation}
where 
\begin{equation}
\mathbf{\mu}_{x,m}=\frac{\underset{n=1}{\overset{N}{\sum}}p\left(m\,|\,\mathbf{y}_{n}\right)\mathbf{x}_{n}}{\underset{n=1}{\overset{N}{\sum}}p\left(m\,|\,\mathbf{y}_{n}\right)},\;\mathbf{\mu}_{y,m}=\frac{\underset{n=1}{\overset{N}{\sum}}p\left(m\,|\,\mathbf{y}_{n}\right)\mathbf{y}_{n}}{\underset{n=1}{\overset{N}{\sum}}p\left(m\,|\,\mathbf{y}_{n}\right)}\label{eq:SPLICEmeans}
\end{equation}
\begin{equation}
\Sigma_{xy,m}=\frac{\underset{n=1}{\overset{N}{\sum}}p\left(m\,|\,\mathbf{y}_{n}\right)\mathbf{x}_{n}\mathbf{y}_{n}^{T}}{\underset{n=1}{\overset{N}{\sum}}p\left(m\,|\,\mathbf{y}_{n}\right)},\;\Sigma_{y,m}=\frac{\underset{n=1}{\overset{N}{\sum}}p\left(m\,|\,\mathbf{y}_{n}\right)\mathbf{y}_{n}\mathbf{y}_{n}^{T}}{\underset{n=1}{\overset{N}{\sum}}p\left(m\,|\,\mathbf{y}_{n}\right)}\label{eq:SPLICEcov}
\end{equation}
i.e., the alignments $p(m\,|\,\mathbf{y}_{n})$ are being used in
place of $p(m\,|\,\mathbf{x}_{n})$ and $p(m\,|\,\mathbf{x}_{n},\mathbf{y}_{n})$
in Eqs. (\ref{eq:SPLICEmeans}) and (\ref{eq:SPLICEcov}) respectively.
Thus from (\ref{eq:mmsee}),
\begin{equation}
\mathbf{A}_{m}=\Sigma_{xy,m}\Sigma_{y,m}^{-1}\label{eq:ammse}
\end{equation}
\begin{equation}
\mathbf{b}_{m}=\mathbf{\mu}_{x,m}-\mathbf{A}_{m}\mathbf{\mu}_{y,m}\label{eq:bmmse}
\end{equation}

To reduce the number of parameters, a simplified model with only bias
$\mathbf{b}_{m}$ is proposed in literature \cite{lideng}.

A diagonal version of Eq. (\ref{eq:mmsee}) can be written as
\begin{equation}
\widehat{x}_{c}=\mu_{x,c}+\frac{\sigma_{xy,c}^{2}}{\sigma_{y,c}^{2}}\left(y-\mu_{y,c}\right)\label{eq:diagsplice}
\end{equation}
where $c$ runs along all components of the features and all mixtures.
Since this method does not capture all the correlations, it suffers
from performance degradation. This shows that noise has significant
effect on feature correlations.

\section{Proposed Modification to SPLICE}

\label{sec:Proposed-Modification}
\begin{figure*}
\centering{}\subfloat[M-SPLICE\label{fig:BD-M-SPLICE}]{\begin{centering}
\includegraphics[scale=0.34]{bd_msplice}
\par\end{centering}

}\quad{}\subfloat[Non-Stereo Method\label{fig:BD-Non-Stereo-Method}]{\begin{centering}
\includegraphics[scale=0.34]{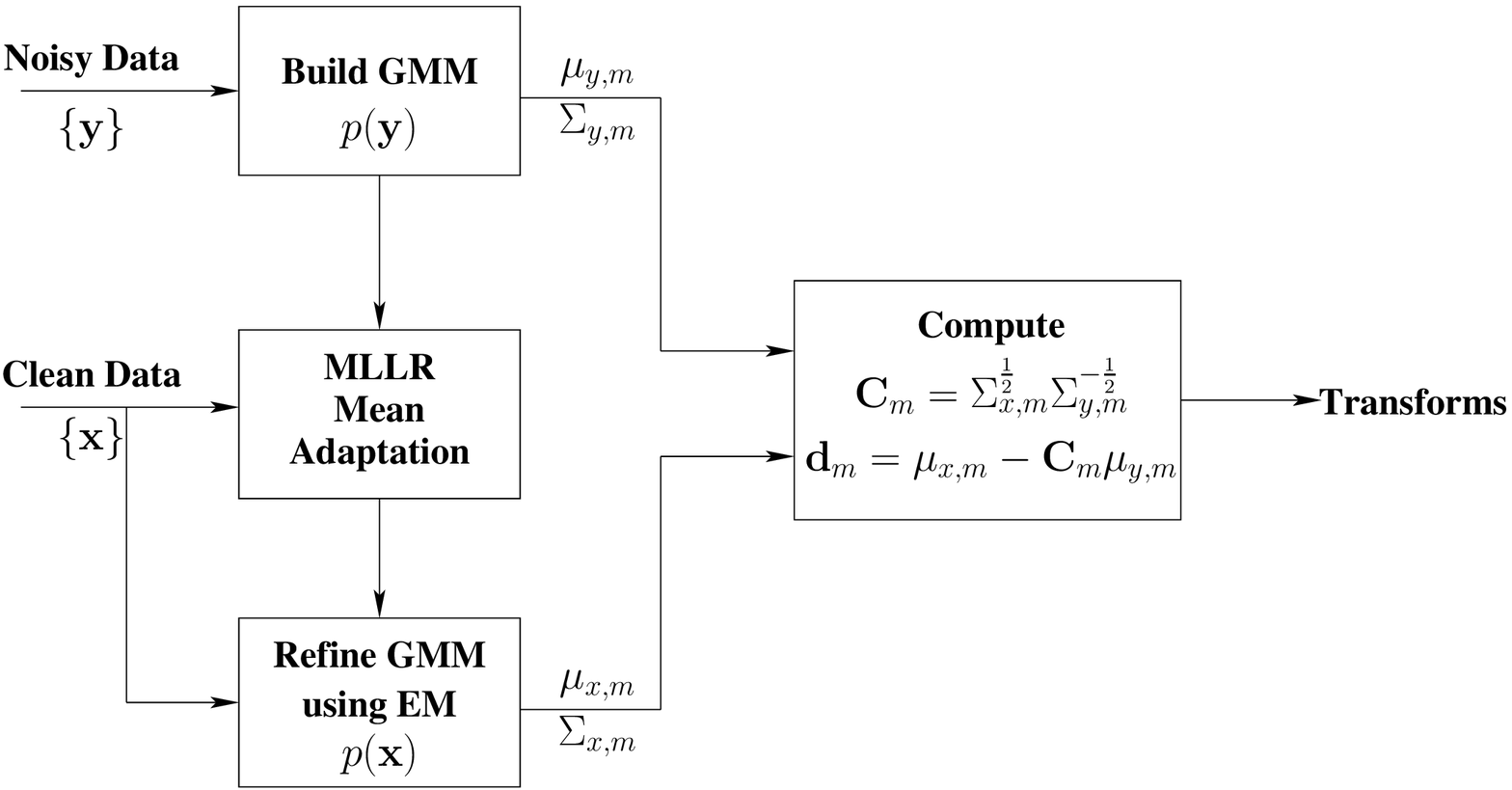}
\par\end{centering}

}\caption{Estimation of piecewise linear transformations\label{fig:BD-Estimation-transforms}}
\end{figure*}

SPLICE assumes that a perfect correlation exists between clean and
noisy stereo features (Eq. (\ref{eq:perfcorr})), which makes the
implementation simple \cite{ssmafify}. But, the actual feature correlations
$\Sigma_{xy,m}$ are used to train SPLICE parameters, as seen in Eq.
(\ref{eq:ammse}). Instead, if the training process also assumes perfect
correlation and eliminates the term $\Sigma_{xy,m}$ during parameter
estimation, it complies with the assumptions and gives improved performance.
This simple modification can be done as follows:

Eq. (\ref{eq:diagsplice}) can be rewritten as
\[
\frac{\widehat{x}-\mu_{x}}{\sigma_{x}}=\frac{\sigma_{xy}^{2}}{\sigma_{x}\sigma_{y}}\left(\frac{y-\mu_{y}}{\sigma_{y}}\right)=\rho\left(\frac{y-\mu_{y}}{\sigma_{y}}\right)
\]
where $\rho=\frac{\sigma_{xy}^{2}}{\sigma_{x}\sigma_{y}}$ is the
correlation coefficient. A perfect correlation implies $\rho=1$.
Since Eq. (\ref{eq:perfcorr}) makes this assumption, we enforce it
in the above equation and obtain
\[
\widehat{x}_{c}=\mu_{x,c}+\frac{\sigma_{x,c}}{\sigma_{y,c}}\left(y-\mu_{y,c}\right)
\]

Similarly, for multidimensional case, the matrix $\Sigma_{x,m}^{-\frac{1}{2}}\Sigma_{xy,m}\Sigma_{y,m}^{-\frac{1}{2}}$
should be enforced to be identity as per the assumption. Thus, we
obtain
\begin{equation}
\widehat{\mathbf{x}}_{m}=\mathbf{\mu}_{x,m}+\Sigma_{x,m}^{\frac{1}{2}}\Sigma_{y,m}^{-\frac{1}{2}}\left(\mathbf{y}-\mathbf{\mu}_{y,m}\right)\label{eq:modsplice}
\end{equation}

Hence M-SPLICE and its updates are defined as 
\begin{equation}
\widehat{\mathbf{x}}=\sum_{m=1}^{M}p\left(m\,|\,\mathbf{y}\right)\left(\mathbf{C}_{m}\mathbf{y}+\mathbf{d}_{m}\right)\label{eq:modsplicex}
\end{equation}
\begin{align}
\mathbf{C}_{m} & =\Sigma_{x,m}^{\frac{1}{2}}\Sigma_{y,m}^{-\frac{1}{2}}\label{eq:cm}\\
\mathbf{d}_{m} & =\mathbf{\mu}_{x,m}-\mathbf{C}_{m}\mathbf{\mu}_{y,m}\label{eq:dm}
\end{align}

All the assumptions of conventional SPLICE are valid for M-SPLICE.
Comparing both the methods, it can be seen from Eqs. (\ref{eq:mmsee})
and (\ref{eq:cm}) that while $\mathbf{A}_{m}$ is obtained using
MMSE estimation framework, $\mathbf{C}_{m}$ is based on whitening
expression. Also, $\mathbf{A}_{m}$ involves cross-covariance term
$\Sigma_{xy,m}$, whereas $\mathbf{C}_{m}$ does not. The bias terms
are computed in the same manner, using their respective transformation
matrices, as seen in Eqs. (\ref{eq:bmmse}) and (\ref{eq:dm}). More
analysis on M-SPLICE is given in Section \ref{sub:Motivation-Non-Stereo}.

\subsection{Training}

The estimation procedure of M-SPLICE transformations is shown in Figure
\ref{fig:BD-M-SPLICE}. The steps are summarised as follows:
\begin{enumerate}
\item Build noisy GMM%
\footnote{We use the term \emph{noisy mixture} to denote a Gaussian mixture
built using noisy data. Similar meanings apply to \emph{clean mixture},\emph{
noisy GMM} and \emph{clean GMM}.%
} $p(\mathbf{y})$ using noisy features $\{\mathbf{y}_{n}\}$ of stereo
data. This gives $\mu_{y,m}$ and $\Sigma_{y,m}$.
\item For every noise frame $\mathbf{y}_{n}$, compute the alignment w.r.t.
the noisy GMM, i.e., $p(m\,|\,\mathbf{y}_{n})$.
\item Using the alignments of stereo counterparts, compute the means $\mu_{x,m}$
and covariance matrices $\Sigma_{x,m}$ of each clean mixture from
clean data $\{\mathbf{x}_{n}\}$.
\item Compute $\mathbf{C}_{m}$ and $\mathbf{d}_{m}$ using Eq. (\ref{eq:cm})
and (\ref{eq:dm}).
\end{enumerate}

\subsection{Testing\label{sub:Testing-M-SPLICE}}

Testing process of M-SPLICE is exactly same as that of conventional
SPLICE, and is summarised as follows:
\begin{enumerate}
\item For each test vector $\mathbf{y}$, compute the alignment w.r.t. the
noisy GMM, i.e., $p(m\,|\,\mathbf{y})$.
\item Compute the cleaned version as:
\[
\widehat{\mathbf{x}}=\sum_{m=1}^{M}p\left(m\,|\,\mathbf{y}\right)\left(\mathbf{C}_{m}\mathbf{y}+\mathbf{d}_{m}\right)
\]

\end{enumerate}

\section{Non-Stereo Extension}

\label{sec:Non-Stereo-Extension}
\begin{figure*}
\noindent \begin{centering}
\subfloat[Separately built noisy and clean GMMs\label{fig:Sky}]{\begin{centering}
\includegraphics[scale=0.36]{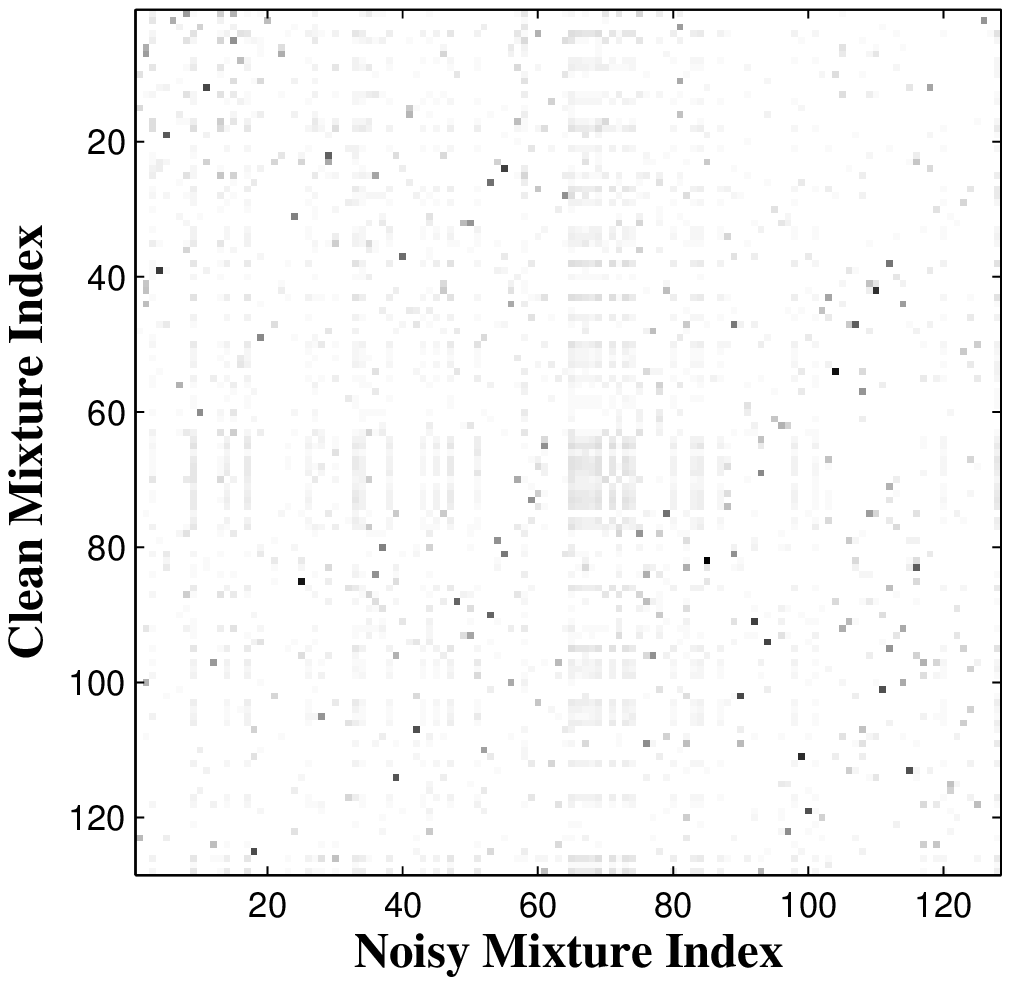}
\par\end{centering}

}\subfloat[GMMs of SPLICE and M-SPLICE\label{fig:SPLICE-M-SPLICE-Sky}]{\begin{centering}
\includegraphics[scale=0.36]{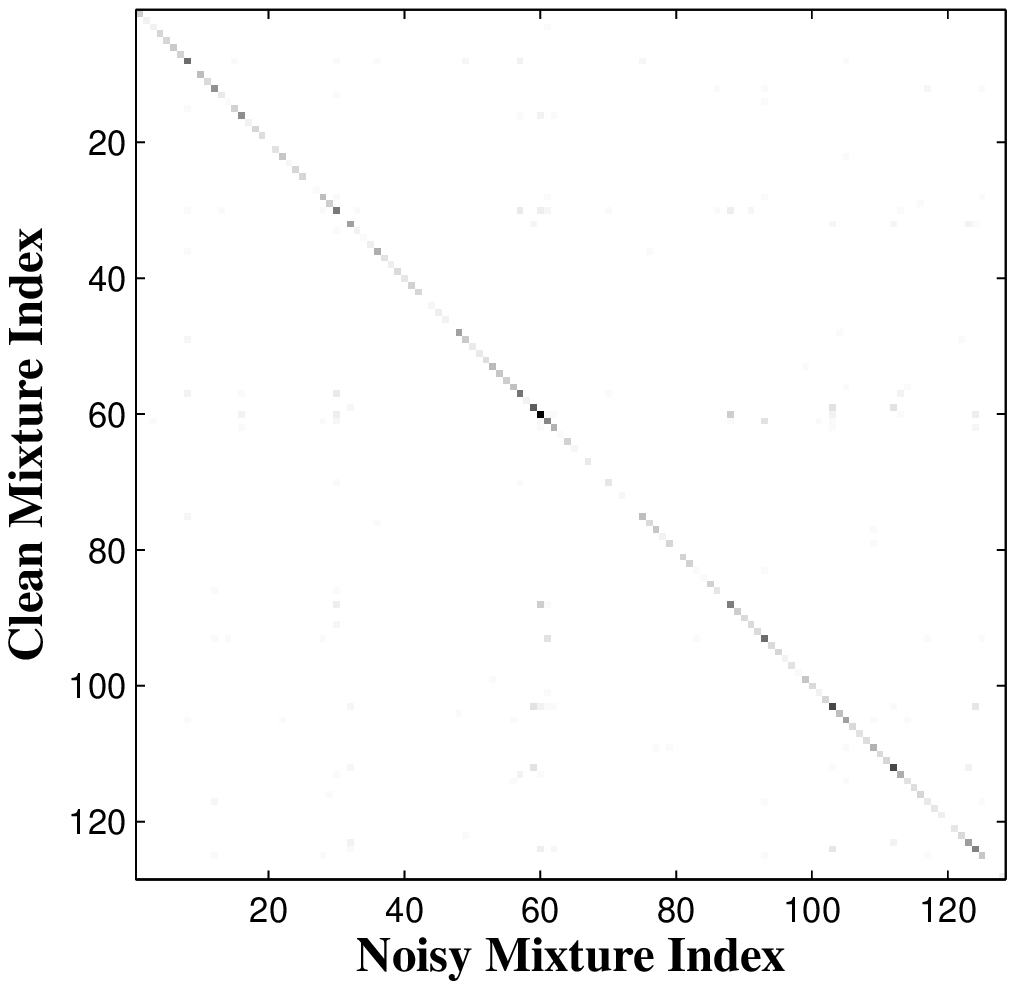}
\par\end{centering}

}\subfloat[Noisy GMM and MLLR-EM based clean GMM\label{fig:Non-Stereo-Sky}]{\begin{centering}
\includegraphics[scale=0.36]{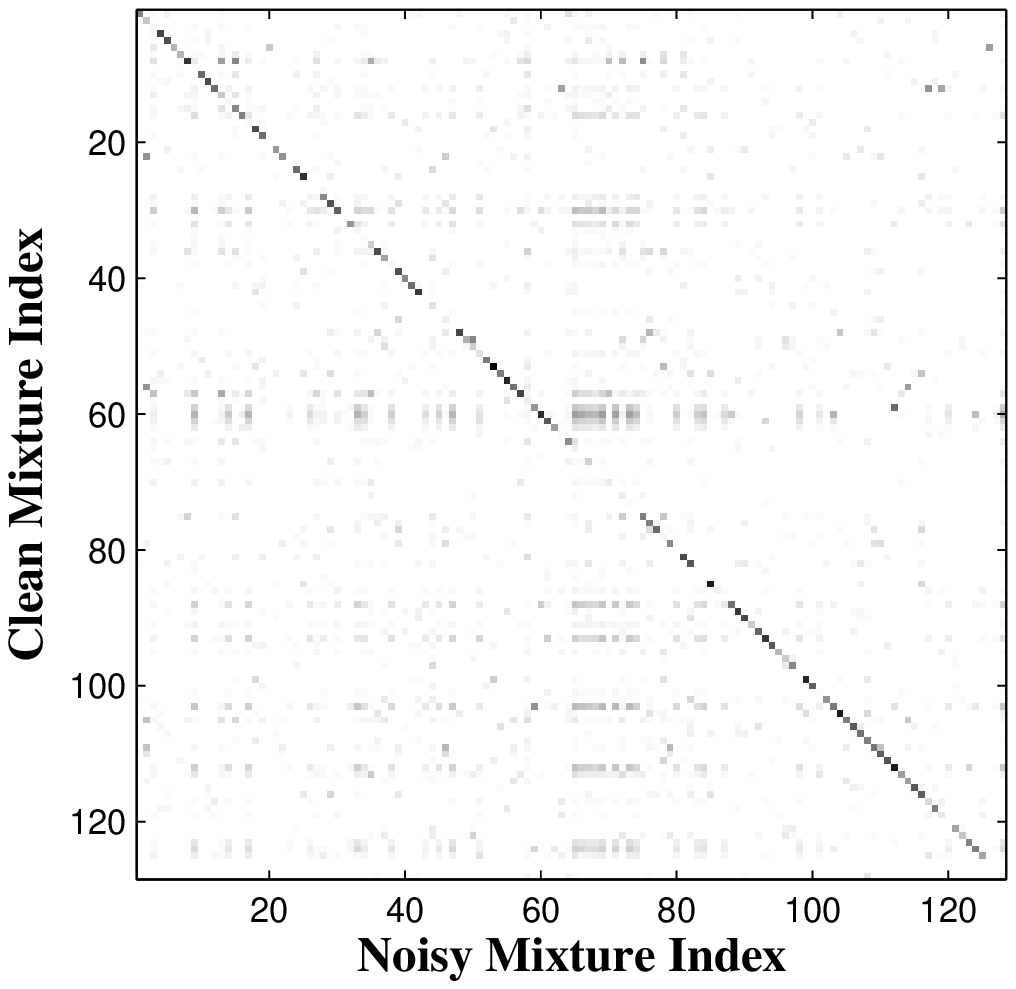}
\par\end{centering}

}
\par\end{centering}

\caption{Mixture assignment distribution plots for Aurora-2 stereo training
data\label{fig:Class-label-distribution-plots}}
\end{figure*}
In this section, we motivate how M-SPLICE can be extended to datasets
which are not stereo recorded. However some noisy training utterances,
which are not necessarily the stereo counterparts of the clean data,
are required.

\subsection{Motivation\label{sub:Motivation-Non-Stereo}}

Consider a stereo dataset of $N$ training frames $(\mathbf{x}_{n},\mathbf{y}_{n})$.
Suppose two $M$ mixture GMMs $p(\mathbf{x})$ and $p(\mathbf{y})$
are independently built using $\{\mathbf{x}_{n}\}$ and $\{\mathbf{y}_{n}\}$
respectively, and each data point is hard-clustered to the mixture
giving the highest probability. We are interested in analysing a matrix
$\mathbf{V}_{M\times M}$, built as 
\[
\mathbf{V}_{ij}=\sum_{n=1}^{N}\mathbbm{1}\left(\mathbf{x}_{n}\in i,\mathbf{y}_{n}\in j\right)
\]
where $\mathbbm{1()}$ is indicator function. In other words, while
parsing the stereo training data, when a stereo pair with clean part
belonging to $i^{th}$ clean mixture and noisy part to $j^{th}$ noisy
mixture is encountered, the $ij^{th}$ element of the matrix is incremented
by unity. Thus each $ij^{th}$ element of the matrix denotes the number
of stereo pairs belong to the $i^{th}$ clean $-$ $j^{th}$ noisy
mixture-pair. When data are soft assigned to all the mixtures, the
matrix can instead be built as:
\[
\mathbf{V}_{ij}=\sum_{n=1}^{N}p(i\,|\,\mathbf{x}_{n})p(j\,|\,\mathbf{y}_{n})
\]

Figure \ref{fig:Sky} visualises such a matrix built using Aurora-2
stereo training data using $128$ mixture models. A dark spot in the
plot represents a higher data count, and a bulk of stereo data points
do belong to that mixture-pair. 

In conventional SPLICE and M-SPLICE, only the noisy GMM $p(\mathbf{y})$
is built, and not $p(\mathbf{x})$. $p\left(m\,|\,\mathbf{y}_{n}\right)$
are computed for every noisy frame, and the same alignments are assumed
for the clean frames $\{\mathbf{x}_{n}\}$ while computing $\mathbf{\mu}_{x,m}$
and \foreignlanguage{english}{$\Sigma_{x,m}$}. Hence $\mathbf{\mu}_{x,m}$,
\foreignlanguage{english}{$\Sigma_{x,m}$} and $p\left(m\,|\,\mathbf{y}\right)$
can be considered as the parameters of a clean hypothetical GMM $p(\mathbf{x})$.
Now, given these GMMs $p(\mathbf{y})$ and $p(\mathbf{x})$, the matrix
$\mathbf{V}$ can be constructed, which is visualised in Figure (\ref{fig:SPLICE-M-SPLICE-Sky}).
Since the alignments are same, and $i^{th}$ clean mixture corresponds
to the $i^{th}$ noisy mixture, a diagonal pattern can be seen.

Thus, under the assumption of Eq. (\ref{eq:perfcorr}), conventional
SPLICE and M-SPLICE are able to estimate transforms from $i^{th}$
noisy mixture to exactly $i^{th}$ clean mixture by maintaining the
mixture-correspondence.

When stereo data is not available, such exact mixture correspondence
do not exist. Figure \ref{fig:Sky} makes this fact evident, since
stereo property was not used while building the two independent GMMs.
However, a sparse structure can be seen, which suggests that for most
noisy mixtures $j$, there exists a unique clean mixture $i^{*}$
having highest mixture-correspondence. This property can be exploited
to estimate piecewise linear transformations from every mixture $j$
of $p(\mathbf{y})$ to a single mixture $i^{*}$ of $p(\mathbf{x})$,
ignoring all other mixtures $i\neq i^{*}$. This is the basis for
the proposed extension to non-stereo data.

\subsection{Implementation}

In the absence of stereo data, our approach is to build two separate
GMMs viz., clean and noisy during training, such that there exists
mixture-to-mixture correspondence between them, as close to Fig. \ref{fig:SPLICE-M-SPLICE-Sky}
as possible. Then whitening based transforms can be estimated from
each noisy mixture to its corresponding clean mixture. This sort of
extension is not obvious in the conventional SPLICE framework, since
it is not straight-forward to compute the cross-covariance terms $\Sigma_{xy,m}$
without using stereo data. Also, M-SPLICE is expected to work better
than SPLICE due to its advantages described earlier.

The training approach of two mixture-corresponded GMMs is as follows:
\begin{enumerate}
\item After building the noisy GMM $p(\mathbf{y})$, it is mean adapted
by estimating a global MLLR transformation using clean training data.
The transformed GMM has the same covariances and weights, and only
means are altered to match the clean data. By this process, the mixture
correspondences are not lost.
\item However, the transformed GMM need not model the clean data accurately.
So a few steps of expectation maximisation (EM) are performed using
clean training data, initialising with the transformed GMM. This adjusts
all the parameters and gives a more accurate representation of the
clean GMM $p(\mathbf{x})$.
\end{enumerate}
Now, the matrix $\mathbf{V}$ obtained through this method using Aurora-2
training data is visualised in Figure \ref{fig:Non-Stereo-Sky}. It
can be noted that no stereo information has been used while obtaining
$p(\mathbf{x})$, following the above mentioned steps, from $p(\mathbf{y})$.
It can be observed that a diagonal pattern is retained, as in the
case of M-SPLICE, though there are some outliers. Since stereo information
is not used, only comparable performances can be achieved. Figure
\ref{fig:BD-Non-Stereo-Method} shows the block diagram of estimating
transformations of non-stereo method. The steps are summarised as
follows:
\begin{enumerate}
\item Build noisy GMM $p(\mathbf{y})$ using noisy features $\{\mathbf{y}\}$.
This gives $\mu_{y,m}$ and $\Sigma_{y,m}$.
\item Adapt the means of noisy GMM $p(\mathbf{y})$ to clean data $\{\mathbf{x}\}$
using global MLLR transformation.
\item Perform at least three EM iterations to refine the adapted GMM using
clean data. This gives $p(\mathbf{x})$, thus $\mu_{x,m}$ and $\Sigma_{x,m}$.
\item Compute $\mathbf{C}_{m}$ and $\mathbf{d}_{m}$ using Eq. (\ref{eq:cm})
and (\ref{eq:dm}).
\end{enumerate}
The testing process is exactly same as that of M-SPLICE, as explained
in Section \ref{sub:Testing-M-SPLICE}.

\section{Additional Run-time Adaptation}

\label{sec:Additional-Run-time-Adaptation}To improve the performance
of the proposed methods during run-time, GMM adaptation to the test
condition can be done in both conventional SPLICE and M-SPLICE frameworks
in a simple manner. Conventional MLLR adaptation on HMMs involves
two-pass recognition, where the transformation matrices are estimated
using the alignments obtained through first pass Viterbi-decoded output,
and a final recognition is performed using the transformed models.

MLLR adaptation can be used to adapt GMMs in the context of SPLICE
and M-SPLICE as follows:
\begin{enumerate}
\item Adapt the noisy GMM through a global MLLR mean transformation
\[
\mathbf{\mu}_{y,m}^{(a)}\leftarrow\mathbf{\mu}_{y,m}
\]

\item Now, adjust the bias term in conventional SPLICE or M-SPLICE as
\begin{equation}
\mathbf{d}_{m}^{(a)}=\mathbf{\mu}_{x,m}-\mathbf{C}_{m}\mathbf{\mu}_{y,m}^{(a)}\label{eq:run-time-adapt}
\end{equation}

\end{enumerate}
This method involves only simple calculation of alignments of the
test data w.r.t. the noisy GMM, and doesn't need Viterbi decoding.
Clean mixture means $\mathbf{\mu}_{x,m}$ computed during training
need to be stored. A separate global MLLR mean transform can be estimated
using test utterances belonging to each noise condition. The steps
for testing process for run-time compensation are summarised as follows:
\begin{table*}[t]
\caption{Results on Aurora-2 Database}

\subfloat[Comparison of SPLICE, M-SPLICE and non-stereo methods\label{tab:Results-on-AURORA-2}]{

\centering{}%
\begin{tabular}{|>{\raggedright}p{1.1cm}|c|c|c|>{\centering}p{1.4cm}|}
\hline 
\textbf{Noise Level} & \textbf{Baseline} & \textbf{SPLICE} & \textbf{M-SPLICE} & \textbf{Non-Stereo Method}\tabularnewline
\hline 
\hline 
Clean & 99.25 & 98.97 & 99.01 & 99.08\tabularnewline
\hline 
SNR 20 & 97.35 & 97.84 & 97.92 & 97.68\tabularnewline
\hline 
SNR 15 & 93.43 & 95.81 & 96.10 & 95.15\tabularnewline
\hline 
SNR 10 & 80.62 & 89.48 & 91.03 & 87.37\tabularnewline
\hline 
SNR 5 & 51.87 & 72.71 & 77.59 & 68.49\tabularnewline
\hline 
SNR 0 & 24.30 & 42.85 & 50.72 & 39.00\tabularnewline
\hline 
SNR -5 & 12.03 & 18.52 & 22.27 & 16.73\tabularnewline
\hline 
\hline 
Test A & 67.45 & 81.39 & 83.47 & 77.44\tabularnewline
\hline 
Test B & 72.26 & 83.24 & 84.18 & 79.63\tabularnewline
\hline 
Test C & 68.14 & 69.42 & \textbf{78.06} & 73.54\tabularnewline
\hline 
\hline 
\emph{Overall} & \emph{69.51} & \emph{79.74} & \textbf{\emph{82.67}} & \emph{77.54}\tabularnewline
\hline 
\end{tabular}}\enskip{}\subfloat[Comparison of adaptation methods\label{tab:Adaptation-Results}]{

\centering{}%
\begin{tabular}{|>{\centering}p{1.2cm}|>{\centering}p{1.6cm}|>{\centering}p{1.7cm}|>{\centering}p{2.7cm}|}
\hline 
\textbf{MLLR (39)} & \textbf{SPLICE + Run-time Adaptation} & \textbf{M-SPLICE + Run-time Adaptation} & \textbf{Non-Stereo Method + Run-time Adaptation}\tabularnewline
\hline 
\hline 
99.28 & 99.05 & 99.02 & 99.08\tabularnewline
\hline 
98.33 & 97.96 & 98.18 & 97.77\tabularnewline
\hline 
96.82 & 96.21 & 96.87 & 95.47\tabularnewline
\hline 
91.88 & 90.61 & 93.10 & 88.80\tabularnewline
\hline 
73.88 & 75.05 & 82.00 & 72.36\tabularnewline
\hline 
41.94 & 46.27 & 57.51 & 44.98\tabularnewline
\hline 
18.71 & 20.10 & 27.32 & 20.43\tabularnewline
\hline 
\hline 
79.31 & 82.45 & 86.47 & 80.12\tabularnewline
\hline 
82.55 & 84.09 & 85.91 & 81.67\tabularnewline
\hline 
79.14 & 73.01 & \textbf{82.90} & 75.79\tabularnewline
\hline 
\hline 
\emph{80.57} & \emph{81.22} & \textbf{\emph{85.53}} & \emph{79.88}\tabularnewline
\hline 
\end{tabular}}
\end{table*}
\begin{table*}
\caption{Results on Aurora-4 Database\label{tab:Results-on-AURORA-4}}

\hspace{1.4cm}%
\begin{tabular}{|c|c|c|c|c|c|c|c|c|c|}
\cline{3-10} 
\multicolumn{1}{c}{} &  & Clean & Car & Babble & Street & Restaurant & Airport & Station & \emph{Average}\tabularnewline
\hline 
\multirow{2}{*}{Baseline} & Mic-1 & 87.63 & 75.58 & 52.77 & 52.83 & 46.53 & 56.38 & 45.30 & \multirow{2}{*}{54.73}\tabularnewline
\cline{2-9} 
 & Mic-2 & 77.40 & 64.39 & 45.15 & 42.03 & 36.26 & 47.69 & 36.32 & \tabularnewline
\hline 
\hline 
\multirow{2}{*}{Non-Stereo Method} & Mic-1 & 86.85 & 77.71 & 62.62 & 58.96 & 55.93 & 61.95 & 55.37 & \multirow{2}{*}{\textbf{\emph{61.66}}}\tabularnewline
\cline{2-9} 
 & Mic-2 & 79.10 & 68.58 & 55.24 & 51.67 & 45.88 & 55.45 & 47.88 & \tabularnewline
\hline 
\end{tabular}
\end{table*}

\begin{enumerate}
\item For all test vectors $\{\mathbf{y}\}$ belonging to a particular environment,
compute the alignments w.r.t. the noisy GMM, i.e., $p(m\,|\,\mathbf{y})$.
\item Estimate a global MLLR mean transformation using $\{\mathbf{y}\}$,
maximising the likelihood w.r.t. $p(\mathbf{y})$.
\item Compute the adapted noisy GMM $p^{(a)}(\mathbf{y})$ using the estimated
MLLR transform. Only the means $\mathbf{\mu}_{y,m}$ of the noisy
GMM would have been adapted as $\mathbf{\mu}_{y,m}^{(a)}$.
\item Using Eq. (\ref{eq:run-time-adapt}), recompute the bias term of SPLICE
or M-SPLICE.
\item Compute the cleaned test vectors as 
\[
\widehat{\mathbf{x}}=\sum_{m=1}^{M}p\left(m\,|\,\mathbf{y}\right)\left(\mathbf{C}_{m}\mathbf{y}+\mathbf{d}_{m}^{(a)}\right)
\]

\end{enumerate}

\section{Experimental Setup}

\label{sec:Experimental-Setup}

Aurora-2 task of 8 kHz sampling frequency \cite{aurora2} has been
used to perform comparative study of the proposed techniques with
the existing ones. Aurora-2 consists of connected spoken digits with
stereo training data. The test set consists of utterances of ten different
environments, each at seven distinct SNR levels. The acoustic word
models for each digit have been built using left to right continuous
density HMMs with 16 states and 3 diagonal covariance Gaussian mixtures
per state. HMM Toolkit (HTK) 3.4.1 has been used for building and
testing the acoustic models.

All SPLICE based linear transformations have been applied on 13 dimensional
MFCCs, including $C_{0}$. During HMM training, the features are appended
with 13 delta and 13 acceleration coefficients to get a composite
39 dimensional vector per frame. Cepstral mean subtraction (CMS) has
been performed in all the experiments. 128 mixture GMMs are built
for all SPLICE based experiments. Run-time noise adaptation in SPLICE
framework is performed on 13 dimensional MFCCs. Data belonging to
each SNR level of a test noise condition has been separately used
to compute the global transformations. In all SPLICE based experiments,
pseudo-cleaning of clean features has been performed.

To test the efficacy of non-stereo method on a database which doesn't
contain stereo data, Aurora-4 task of 8 kHz sampling frequency has
been used. Aurora-4 is a continuous speech recognition task with clean
and noisy training utterances (non-stereo) and test utterances of
14 environments. Aurora-4 acoustic models are built using crossword
triphone HMMs of 3 states and 6 mixtures per state. Standard WSJ0
bigram language model has been used during decoding of Aurora-4. Noisy
GMM of 512 mixtures is built for evaluating non-stereo method, using
7138 utterances taken from both clean and multi-training data. This
GMM is adapted to standard clean training set to get the clean GMM.

\subsection{Results}

Tables \ref{tab:Results-on-AURORA-2} and \ref{tab:Adaptation-Results}
summarise the results of various algorithms discussed, on Aurora-2
dataset. All the results are shown in \% accuracy. All SNRs levels
mentioned are in decibels. The first seven rows report the overall
results on all 10 test noise conditions. The rest of the rows report
the average values in the SNR range 20-0 dB. Table \ref{tab:Results-on-AURORA-4}
shows the experimental results on Aurora-4 database.

For reference, the result of standard MLLR adaptation on HMMs \cite{mllrgales}
has been shown in Table \ref{tab:Adaptation-Results}, which computes
a global 39 dimensional mean transformation, and uses two-pass Viterbi
decoding.

It can be seen that M-SPLICE improves over SPLICE at all noise conditions
and SNR levels and gives an \emph{absolute} improvement of $8.6\%$
in test-set C and $2.93\%$ overall. Run-time compensation in SPLICE
framework gives improvements over standard MLLR in test-sets A and
B, whereas M-SPLICE gives improvements in all conditions. Here $9.89\%$
absolute improvement can be observed over SPLICE with run-time noise
adaptation, and $4.96\%$ over standard MLLR. Finally, non-stereo
method, though not using stereo data, shows $10.37\%$ and $6.93\%$
absolute improvements over Aurora-2 and Aurora-4 baseline models respectively,
and a slight degradation w.r.t. SPLICE in all test cases. Run-time
noise adaptation results of non-stereo method are comparable to that
of standard MLLR, and are computationally less expensive.

\section{Discussion}

\label{sec:Discussion}

In terms of computational cost, the methods M-SPLICE and non-stereo
methods are identical during testing as compared to conventional SPLICE.
Also, there is almost negligible increase in cost during training.
The MLLR mean adaptation in both non-stereo method and run-time adaptation
are computationally very efficient, and do not need Viterbi decoding.

In terms of performance, M-SPLICE is able to achieve good results
in all cases without any use of adaptation data, especially in unseen
cases. In non-stereo method, one-to-one mixture correspondence between
noise and clean GMMs is assumed. The method gives slight degradation
in performance. This could be attributed to neglecting the outlier
data.

Comparing with other existing feature normalisation techniques, the
techniques in SPLICE framework operate on individual feature vectors,
and no estimation of parameters is required from test data. So these
methods do not suffer from test data insufficiency problems, and are
advantageous for shorter utterances. Also, the testing process is
usually faster, and are easily implementable in real-time applications.
So by extending the methods to non-stereo data, we believe that they
become more useful in many applications.

\section{Conclusion and Future Work}

\label{sec:Conclusion-and-Future}

A modified version of the SPLICE algorithm has been proposed for noise
robust speech recognition. It is better compliant with the assumptions
of SPLICE, and improves the recognition in highly mismatched and unseen
noise conditions. An extension of the methods to non-stereo data has
been presented. Finally, a convenient run-time adaptation framework
has been explained, which is computationally much cheaper than standard
MLLR on HMMs. In future, we would like to improve the efficiency of
non-stereo extensions of SPLICE, and extend M-SPLICE in uncertainty
decoding framework.

\bibliographystyle{ieeetr}
\bibliography{SPLICE_ASRU}

\end{document}